\documentclass[letterpaper, 10 pt, conference]{ieeeconf}  

\IEEEoverridecommandlockouts                              

\usepackage{graphicx}
\usepackage{amsmath,amssymb,amsfonts}
\usepackage[caption=false]{subfig}
\usepackage[ruled,linesnumbered, noend]{algorithm2e}
\usepackage{dirtytalk}
\usepackage{hyperref}
\usepackage{gensymb}
\usepackage{tabularx}
\usepackage{multirow}
\usepackage[short]{optidef}
\usepackage{booktabs}
\usepackage{siunitx}

\makeatletter
\newcommand{\removelatexerror}{\let\@latex@error\@gobble}
\makeatother
\SetKwComment{Comment}{$\triangleright$\ }{}

\title{\LARGE \bf
DPMPC-Planner: A real-time UAV trajectory planning framework for complex static environments with dynamic obstacles}

\author{Zhefan Xu, Di Deng, Yiping Dong, and Kenji Shimada 
\thanks{Zhefan Xu, Di Deng, Yiping Dong, and Kenji Shimada are with Department of Mechanical Engineering, Carnegie Mellon University, 5000 Forbes Ave, Pittsburgh, PA, 15213, USA.
        {\tt\small zhefanx@andrew.cmu.edu}}%
}

\begin{document}

\maketitle
\thispagestyle{empty}
\pagestyle{empty}

\begin{abstract}

Safe UAV navigation is challenging due to the complex environment structures, dynamic obstacles, and uncertainties from measurement noises and unpredictable moving obstacle behaviors. Although plenty of recent works achieve safe navigation in complex static environments with sophisticated mapping algorithms, such as occupancy map and ESDF map, these methods cannot reliably handle dynamic environments due to the mapping limitation from moving obstacles. To address the limitation, this paper proposes a trajectory planning framework to achieve safe navigation considering complex static environments with dynamic obstacles. To reliably handle dynamic obstacles, we divide the environment representation into static mapping and dynamic object representation, which can be obtained from computer vision methods. Our framework first generates a static trajectory based on the proposed iterative corridor shrinking algorithm. Then, reactive chance-constrained model predictive control with temporal goal tracking is applied to avoid dynamic obstacles with uncertainties. The simulation results in various environments demonstrate the ability of our algorithm to navigate safely in complex static environments with dynamic obstacles.

\end{abstract}

\section{Introduction}
\begin{figure}[t]
    \vspace{0.2cm}
    \centering
    \includegraphics[scale=0.17]{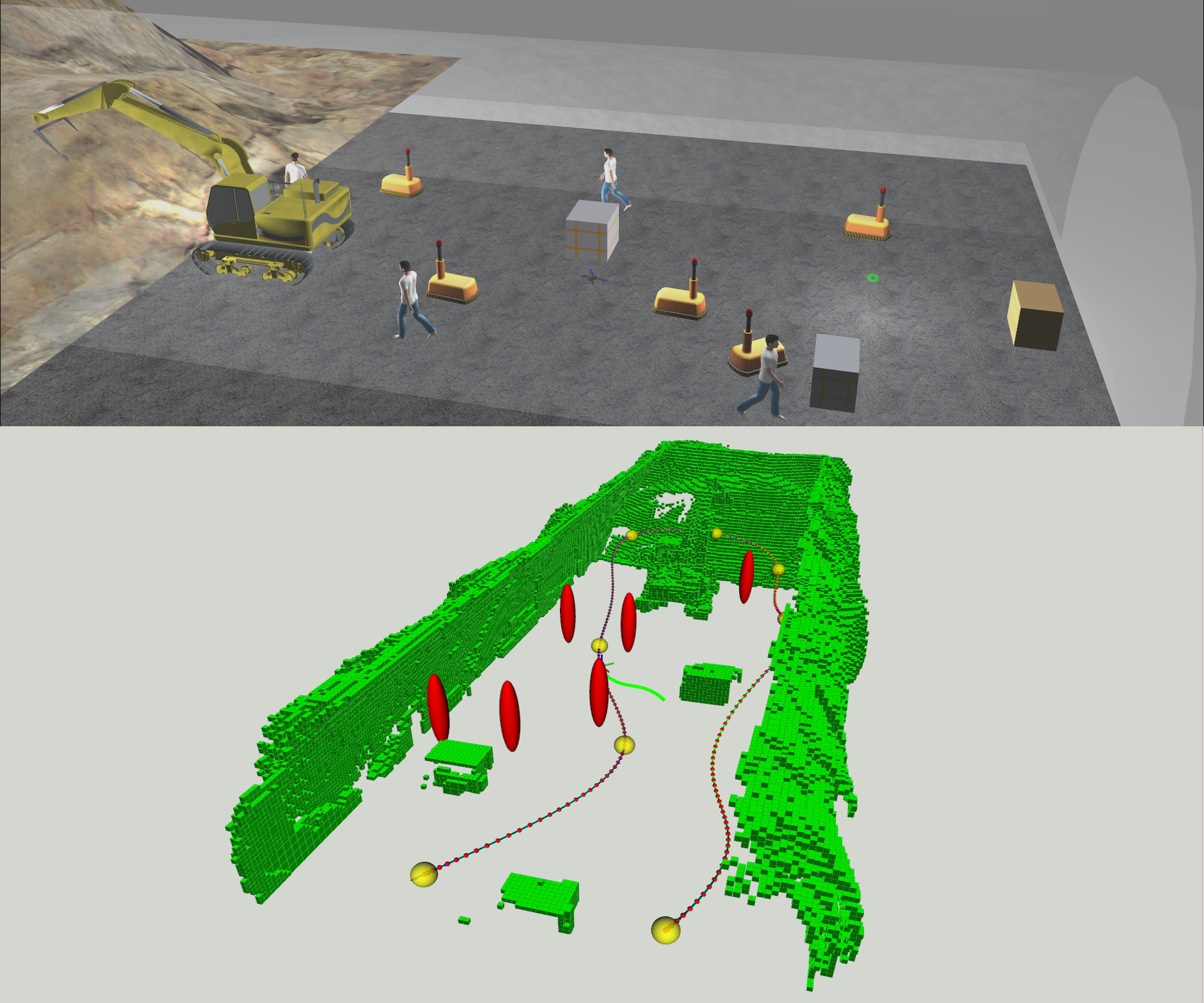}
    \caption{The proposed method running in the Tunnel environment with walking persons and moving robots. The bottom figure shows how the quadrotor avoids dynamic obstacles based on the static trajectory. }
    \label{intro_figure}
\end{figure}

With the increasing usage of autonomous UAVs in industries, online trajectory generation becomes crucial for safety and autonomy in a complex structured environment with moving humans and robots, as shown in Fig. \ref{intro_figure}. In these scenarios, robots are required to navigate to their goals in cluttered environments and guarantee safety with humans. Consequently, a safe trajectory planning framework is essential to deal with complex environment structures, dynamic obstacles, and uncertainties from measurement noise and unpredictable moving obstacle behaviors. 

There are two main challenges of safe trajectory planning in dynamic environments, making the problem not perfectly solved by past research. Firstly, the complexity of the environments makes optimal trajectory generation computationally expensive considering vehicle dynamics. Even though some works \cite{9422918}\cite{9309347}\cite{8206119}\cite{7138978} prove the real-time performance in static environment navigation, their methods only depend on specific map representation, such as occupancy map \cite{hornung13auro} or ESDF map \cite{8202315}\cite{8968199}, which cannot reliably encode dynamic obstacles. Secondly, the algorithm needs to run fast enough considering both static and dynamic obstacles. Previous publications \cite{zhu2019chance}\cite{9197481}\cite{castillo2020real} provide a reliable solution based on the model predictive control and using geometric representation for modeling dynamic obstacles. However, complex static environment structures are not considered for trajectory generation. 

In this paper, we present a novel trajectory planning framework named \textbf{D}ynamic \textbf{P}olynomial-based \textbf{M}odel \textbf{P}redictive \textbf{C}ontrol (\textbf{DPMPC}) planner. This framework contains two planning layers to deal with static environments and dynamic obstacles, which fully utilizes both occupancy map and geometric obstacle representation. In the static layer, we adopt the polynomial-based minimum snap trajectory optimization scheme with our iterative corridor shrinking algorithm to find the static trajectory. Then, reactive online trajectories are generated by the dynamic layer using chance-constrained model predictive control, and dynamic obstacles are avoided by our temporal goal tracking method. Simulation experiments demonstrate that our method can help UAVs safely navigate in cluttered dynamic environments. The novelties and contributions of this work are:
\begin{itemize}
  \item \textbf{Uncertainty-aware trajectory planner:} This work presents a two-layer DPMPC planner for safe trajectory generation in complex static environments with dynamic obstacles with uncertainties. 
  \item \textbf{Temporal goal tracking with MPC:} The temporal goal tracking achieves dynamic obstacle avoidance with chance-constrained model predictive control to handle measurement and obstacle uncertainties.
  \item \textbf{Iterative corridor shrinking algorithm:} The proposed iterative corridor shrinking algorithm provides a computationally efficient method to generate static trajectories.

  
\end{itemize}
\section{Related Work}
Trajectory generation in static environments can be viewed as a constrained optimization problem. In \cite{5980409}, the differential flatness of quadrotors shows the effectiveness of minimum snap optimization. Based on that, \cite{6225009} and \cite{7138978} apply mixed integer programming to avoid static obstacles. \cite{richter2016polynomial} adopts a similar scheme but ensures trajectory to be collision-free by adding intermediate waypoints. Inspired by \cite{5152817}\cite{5980280}, some works make the optimization problem unconstrained. \cite{9422918} formulates unconstrained optimization with ESDF map to achieve aggressive flight with unknown static obstacles, and \cite{9309347} furthermore saves computation by reducing the dependency on ESDF. Besides, \cite{8206119} provides a search-based method to avoid computational complexity from the optimal control problem. In the UAV exploration, \cite{9362184} applies an incremental sampling method to shorten the replanning time for obstacle avoidance. Although the above methods prove successful navigation in complex environments, most algorithms rely on either occupancy or ESDF map, making it hard to encode moving obstacles reliably.

Dynamic obstacle avoidance has been investigated in early works and has remained open in recent years. \cite{khatib1986real} first proposes artificial potential field to avoid obstacles, and \cite{fiorini1998motion} defines velocity obstacle that suggests picking non-collision velocities. In UAV planning, obstacle avoidance by model predictive control (MPC) has gained great attraction in recent years. In \cite{shim2003decentralized}, obstacle potential field cost \cite{khatib1986real} is added to MPC for inter-robot collision avoidance. To account for uncertainties, \cite{blackmore2011chance} adopts the disjunctive programming to solve the chance-constrained MPC, but it is still computationally intensive. Based on \cite{shim2003decentralized} and \cite{blackmore2011chance}, \cite{zhu2019chance} presents an online linearized chance-constrained MPC to deal with dynamic obstacles with real-time performance. Compared to the deterministic constraint in \cite{kamel2017robust}, \cite{zhu2019chance} makes the trajectory less conservative. Similar to \cite{zhu2019chance}, \cite{castillo2020real} achieves online avoidance by adding bounds on disjunctive constraints in their chance-constrained MPC. \cite{9197481} extends the idea of \cite{zhu2019chance} by using UAV's onboard vision. However, these methods apply geometric and analytical formulas for obstacle representation, which is insufficient to describe environments' complex structures compared to the occupancy map and can make robots fail to navigate in cluttered environments. Unlike previous works, our proposed method combines the advantages of static mapping (e.g., occupancy map) and geometric representation of obstacles to generate trajectories in complex structured environments with online dynamic obstacle avoidance. 


\section{Methodology}
\begin{figure}[t]
    \centering
    \includegraphics[scale=0.262]{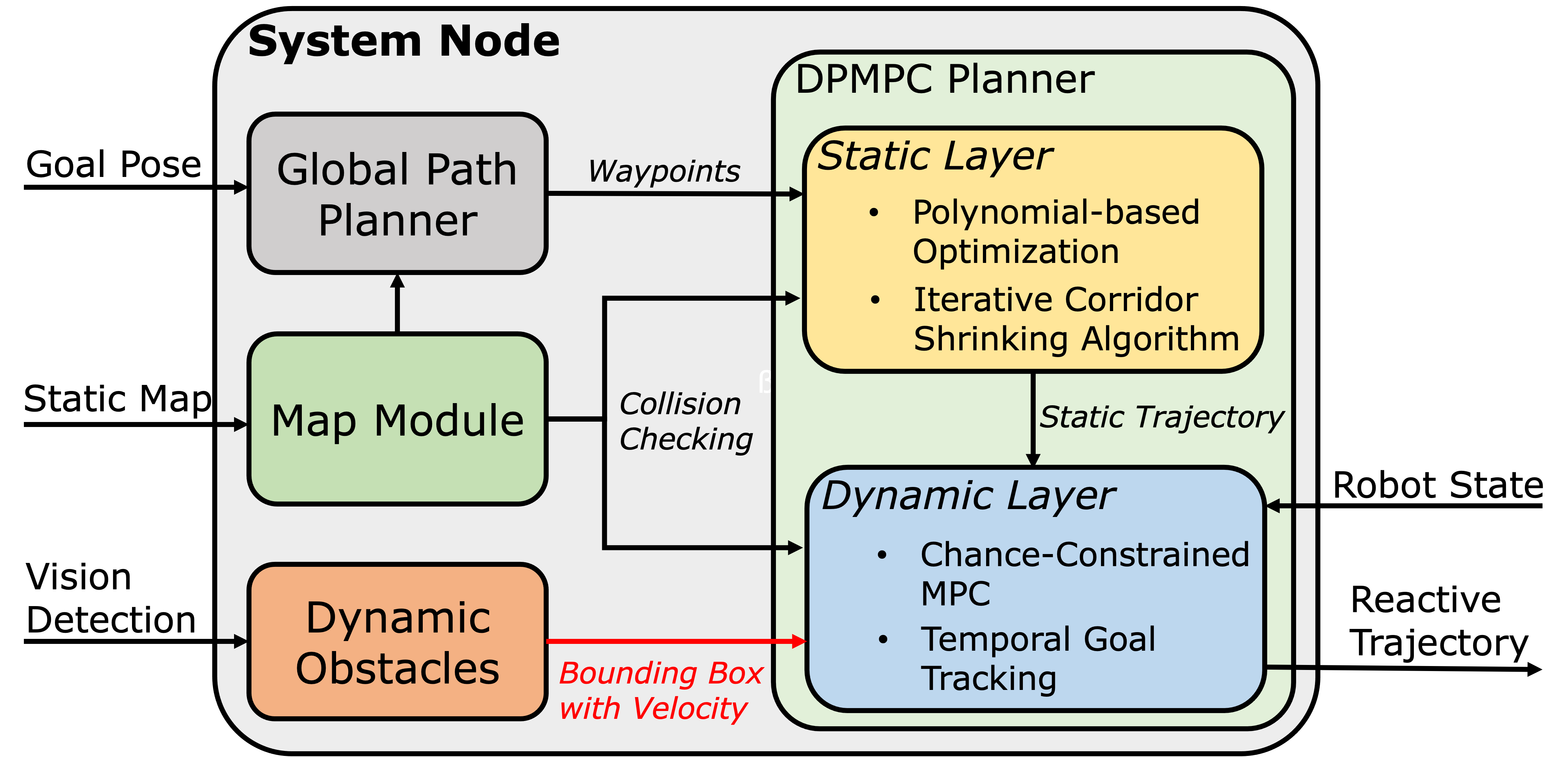}
    \caption{System Overview. The static layer applies polynomial-based optimization and iterative corridor shrinking algorithm to generate the static trajectory. Then, the dynamic layer uses chance-constrained MPC with temporal goal tracking for obstacle avoidance.}
    \label{system_overview}
\end{figure}


\subsection{System Framework}
The system mainly has four parts: global path planner, map module, dynamic obstacle representation, and the proposed DPMPC planner as shown in Fig. \ref{system_overview}. At the first stage, the global path planner takes the goal as input to generate a high-level collision-free waypoint path. Then, the DPMPC planner takes these waypoints into the static layer to optimize a static trajectory. After that, the static trajectory will be fed into the dynamic layer yielding the final reactive trajectory for navigation and obstacle avoidance. Besides the static trajectory, the dynamic layer also takes the dynamic obstacle information, which contains obstacle poses with bounding boxes and estimated velocities, into its MPC constraints. Throughout the process, the map module provides the collision checking function for each planning layer.

\subsection{Static Layer}
The static layer generates the collision-free static trajectory based on the proposed iterative corridor shrinking algorithm as shown in Alg.\ref{ICS}, which relies on the polynomial-based minimum-snap trajectory optimization.  In Line \ref{corridor_size}, we define a corridor bounding box size $\mathcal{C}_{\text{box}}$ for each sample position on the trajectory. Note that we do not need to guarantee the corridor bounding box to be collision-free but only pick a reasonable initial size (e.g., 0.5m). The sample positions are obtained by dividing the continuous trajectory at $\Delta T$ resolution. The main loop (Lines \ref{loop_start}-\ref{loop_end}) will iterate until a collision-free trajectory is generated. Inside the loop, we first initialize the polynomial-based optimization, which will be discussed later in this section. Then, the corridor bounding boxes are added as linear inequality constraints to the optimization problem. After that, we use the map module to check the collision of the optimized trajectory. Finally, the corridor bounding box of the trajectory shrinks by user-defined shrink factor $\mathcal{\alpha}$ (Line \ref{shrink_factor}), and this process repeats until a feasible trajectory is found. Note that the shrinking factor is a tuning parameter typically range from 0.5 to 1.

\begin{algorithm}[t]
\caption{Iterative Corridor Shrinking}
\SetAlgoNoLine%
\label{ICS}
 $\mathcal{M} \gets \text{Static Map}$\;
 $\mathcal{C}_{\text{box}} \gets \text{Corridor Bounding Box Size}$\; \label{corridor_size}
 $\mathcal{V}_{\text{d}} \gets \text{Desired Velocity}$\;
 $\mathcal{\alpha} \gets \text{0.9}$ \Comment*[r]{user-defined shrink factor} \label{shrink_factor}
 $\mathcal{T}_{\text{collision}} \gets true$ \Comment*[r]{trajectory collision}
 
 \While{$\mathcal{T}_{\text{collision}}$}{ \label{loop_start}
    $\mathcal{P}_{\text{opt}} \gets \textbf{initPolyOptimization}(\mathcal{V}_{\text{d}})$\;
    $\mathcal{P}_{\text{opt}}.\textbf{addCorridorConstraints}(\mathcal{C}_{\text{box}}, \Delta T)$\;
    $\mathcal{\sigma}_{\text{traj}} \gets \mathcal{P}_{\text{opt}}.\textbf{solve}()$\;
    $\mathcal{T}_{\text{collision}} \gets \mathcal{M}.\textbf{checkCollision}(\mathcal{\sigma}_{\text{traj}}, \Delta T)$\;
    $\mathcal{C}_{\text{box}} \gets \mathcal{\alpha} \cdot \mathcal{C}_{\text{box}}$\; \label{loop_end}
 }
$\textbf{return} \ \mathcal{\sigma}_{\text{traj}}$\;
\end{algorithm}

The polynomial-based optimization minimizes the snap of the trajectory \cite{5980409} \cite{richter2016polynomial}, which is the fourth derivative of the position. Assuming the total number of waypoints is $N+1$, the whole trajectory is constructed by piecewise-continuous polynomial segments crossing nearby waypoints:
\begin{equation}
\sigma_{\text{traj}}(t) = \{^{1}\sigma_{\text{traj}}(t), ^{2}\sigma_{\text{traj}}(t) \dots ^{N}\sigma_{\text{traj}}(t)\}.
\end{equation}
Based on \cite{5980409}, we represent each segment of trajectory between two waypoints using polynomials as:
\begin{equation}
^{n}\sigma_{\text{traj}}(t)= 
\begin{bmatrix}
^{n}x(t)\\
^{n}y(t)\\
^{n}z(t)
\end{bmatrix} = 
\begin{bmatrix}
\sum_{i=0}^{d}{^{n}w_{i,x} t^{i}}\\
\sum_{i=0}^{d}{^{n}w_{i,y} t^{i}} \\
\sum_{i=0}^{d}{^{n}w_{i,z} t^{i}}
\end{bmatrix}, t \in [t_{n}, t_{n+1}],
\\
\end{equation}
where $t_{n}$ represents the time of arriving at the $n$th waypoint, and $d$ is the degree of the polynomials. So, the optimization problem can be written as the following quadratic programming (QP):
\begin{mini!}[2]
{\textbf{w}}{F(\textbf{w}) = \int_{t_{0}}^{t_{N}} \begin{Vmatrix} \frac{d^{k} \sigma_{\text{traj}}}{dt^{k}} \end{Vmatrix}^{2}\,dt + \eta \begin{Vmatrix} \textbf{w} \end{Vmatrix}^2 }{}{} \label{objective}
\addConstraint{^{n}\sigma_{\text{traj}}(t_{n+i}) }{=^{n+i}P_{\text{wp}}, i = 0, 1}{} \label{connect_waypoint}
\addConstraint{\frac{^{n-1}d^{m} {\sigma_{\text{traj}}} (t_{n})}{dt^{m}}}{=\frac{^{n}d^{m} {\sigma_{\text{traj}}} (t_{n})}{dt^{m}}}{} \label{continuity}
\addConstraint{\forall m \in \{1, \dots, k-1\}}{,\forall n \in \{1,\dots,N \}}{} \label{all_points}
\addConstraint{P_{\text{ip}} - \mathcal{C}_{\text{box} \leq \sigma_{\text{traj}}(t_{s})}}{  \leq P_{\text{ip}} + \mathcal{C}_{\text{box}}}{} \label{collision}
\addConstraint{\forall t_{s} \in \{t_{0}, t_{0}+\Delta T, \dots, t_{N}\}} \label{time_sample}
\end{mini!}

\noindent In the objective function (Eqn. \ref{objective}),  \textbf{w} is the coefficient vector of all polynomials, $k$ is 4 to minimize the trajectory snap. Note that we add a regularization term to increase the stability of the optimization. The constraints from Eqn. \ref{connect_waypoint}-\ref{all_points} ensure the trajectory to pass through all waypoints and maintain continuity at connecting points, where $P_{\text{wp}}$ is the position of the waypoint. For the constraints from Eqn. \ref{collision}-\ref{time_sample}, they discretize the trajectory at $\Delta T$ sample resolution (Eqn. \ref{time_sample}) and apply bounding box constraints to each trajectory sample position with respect to the interpolation position $P_{\text{ip}}$ on the line segment between two nearby waypoints (Eqn. \ref{collision}). Even though the iterative corridor shrinking algorithm requires restarting optimization multiple times, this QP formulation helps the algorithm run in real-time.

\subsection{Dynamic Layer}
In this section, we first introduce the chance-constrained model predictive control formulation in our dynamic layer. Then, the definition of avoidance target is introduced, and we present the temporal goal tracking method for dynamic obstacle avoidance based on the avoidance target selection.

\textbf{Chance-constrained MPC formulation:} There are two requirements for our model predictive control: (a) tracking the static trajectory and (b) avoiding dynamic obstacles. To deal with uncertainties from localization and obstacle perception, we formulate the chance-constrained problem based on \cite{zhu2019chance} as following:
\begin{mini!}[2]
    {\textbf{x}^{1:N}, \textbf{u}^{1:N}}{\sum_{k=1}^{N-1} {\begin{Vmatrix} \textbf{x}^{k} - \textbf{x}^{k}_{\text{ref}} \end{Vmatrix}^2 + \begin{Vmatrix} \textbf{u}^{k} \end{Vmatrix}^2}}{}{} \label{mpc_objective}
\addConstraint{\textbf{x}^{0}}{=\textbf{x}(t_{0})}{}
\addConstraint{\textbf{x}^{k}}{=f(\textbf{x}^{k-1}, \textbf{u}^{k-1})}{} \label{dynamic_model}
\addConstraint{\text{Pr}(\textbf{x}^{k} \in C_{i})}{\geq \delta, \forall i \in I_{o}}{} \label{chance_constrained}
\addConstraint{\textbf{u}_{\text{min}} \leq}{\textbf{u} \leq  \textbf{u}_{\text{max}}}{}
\addConstraint{\forall k \in \{1, \ldots , N\}}
\end{mini!}
\noindent In the objective function (Eqn. \ref{mpc_objective}), the reference trajectory is the static trajectory when the robot does not meet obstacles, and the robot will track the static trajectory in this case. When the robot meets obstacles, the reference trajectory is switched to the temporal goal, as we will discuss in the later section to avoid obstacles. The dynamics model is represented in Eqn. \ref{dynamic_model}, which we refer to \cite{5980409}\cite{kamel2017model} for further details.

The collision constraint is applied using chance-constrained scheme to deal with uncertainties (Eqn. \ref{chance_constrained}). We first assume the positions of robots and obstacles follow the Gaussian distribution as $\textbf{p}_{r} \sim \mathcal{N}(\overline{\textbf{p}}_{r}, \Sigma_{r})$ and $\textbf{p}_{o} \sim \mathcal{N}(\overline{\textbf{p}}_{o}, \Sigma_{o})$, respectively. The collision geometry of the robot is represented by a sphere with radius $r$, and for obstacles, we use the smallest ellipsoid enclosing its bounding box, parametrized by semi-axis lengths $a$, $b$, and $c$. So, the set of states colliding with the $i$-th obstacle at the $k$th step can be written in the following deterministic form:
\begin{equation}
C_{i}^{k} = \{\textbf{x}^{k} | \begin{Vmatrix} \textbf{p}_{r}^{k} - \textbf{p}_{o}^{k}  \end{Vmatrix}_{\text{Q}_{c}} \leq 1\}, \label{deterministic_collision}
\end{equation} 
where $\text{Q}_{c}$ is $diag(\frac{1}{{r+a}^2}, \frac{1}{{r+b}^2}, \frac{1}{{r+c}^2})$. With the Gaussian assumption and the definition of collision condition (Eqn. \ref{deterministic_collision}), we can further derive the probability of collision for the $i$-th obstacle at the $k$th step as:
\begin{equation}
\text{Pr}(\textbf{x}^{k} \in C_{i}^{k}) = \int_{\begin{Vmatrix} \textbf{p}_{r}^{k} - \textbf{p}_{o}^{k}  \end{Vmatrix}_{\text{Q}_{c}} \leq 1} p(\textbf{p}_{r}^{k} - \textbf{p}_{o}^{k})\,d(\textbf{p}_{r}^{k} - \textbf{p}_{o}^{k}), \label{gaussion_integral}
\end{equation}
This equation requires the integral of a Gaussian distribution over an ellipsoid which does not have an analytical solution. So, we first perform a coordinate transformation to make the ellipsoid a sphere so as to eliminate $\text{Q}_{c}$: 
\begin{equation}
C_{i}^{k} = \{\textbf{x}^{k} | \begin{Vmatrix} \textbf{p}_{r, t}^{k} - \textbf{p}_{o, t}^{k}  \end{Vmatrix} \leq 1\}, \label{linearized_equation}
\end{equation} 
where $\textbf{p}_{r, t}^{k} = \text{Q}_{c}^{\frac{1}{2}} \textbf{p}_{r}^{k}$, $\textbf{p}_{o, t}^{k} = \text{Q}_{c}^{\frac{1}{2}} \textbf{p}_{o}^{k}$, and their corresponding covariance matrices are: $\Sigma_{r, t}^{k} = \text{Q}_{c}^{\frac{1}{2}T} \Sigma_{r}^{k} \text{Q}_{c}^{\frac{1}{2}}$ and $\Sigma_{o, t}^{k} = \text{Q}_{c}^{\frac{1}{2}T} \Sigma_{o}^{k} \text{Q}_{c}^{\frac{1}{2}}$, respectively. Then, we approximate Eqn. \ref{deterministic_collision} using linearization as proposed in \cite{zhu2019chance}: 
\begin{equation}
C_{i, \text{approx}}^{k} = \{\textbf{x}^{k} |  \textbf{a}^{kT}(\textbf{p}_{r, t}^{k} - \textbf{p}_{o, t}^{k}) \leq 1\}, \label{linearized_equation}
\end{equation}
where $\textbf{a}^{k} = \frac{\overline{\textbf{p}}_{r, t}^{k} - \overline{\textbf{p}}_{o, t}^{k}}{\begin{Vmatrix} \overline{\textbf{p}}_{r, t}^{k} - \overline{\textbf{p}}_{o, t}^{k} \end{Vmatrix}}$. Based on the linearized collision condition, Eqn. \ref{chance_constrained} can be written in the following form:
\begin{equation}
\text{Pr}(\textbf{x}^{k} \in C_{i, \text{approx}}^{k}) = \text{Pr}(\textbf{a}^{kT}(\textbf{p}_{r, t}^{k} - \textbf{p}_{o, t}^{k}) \leq 1) \leq \delta, 
\end{equation}
and we can further derive the deterministic form as:
\begin{equation}
\begin{split}
\textbf{a}^{kT}(\textbf{p}_{r, t}^{k} - \textbf{p}_{o, t}^{k}) - 1 \geq \text{erf}^{-1}
(1-2\delta)\\
\sqrt{2\textbf{a}^{k}(\Sigma_{r, t}^{k}+\Sigma_{o, t}^{k})\textbf{a}^{k}}, 
\end{split}
\end{equation}
which provides an analytical expression for Eqn. \ref{chance_constrained} and helps us solve the optimal control problem in real-time.

\textbf{Temporal Goal Tracking:}
As the robot approaches obstacles, we need to select the avoidance target, a special position on the static trajectory, as the detour goal for collision avoidance. 

\begin{figure}[t]
    \centering
    \includegraphics[scale=0.44]{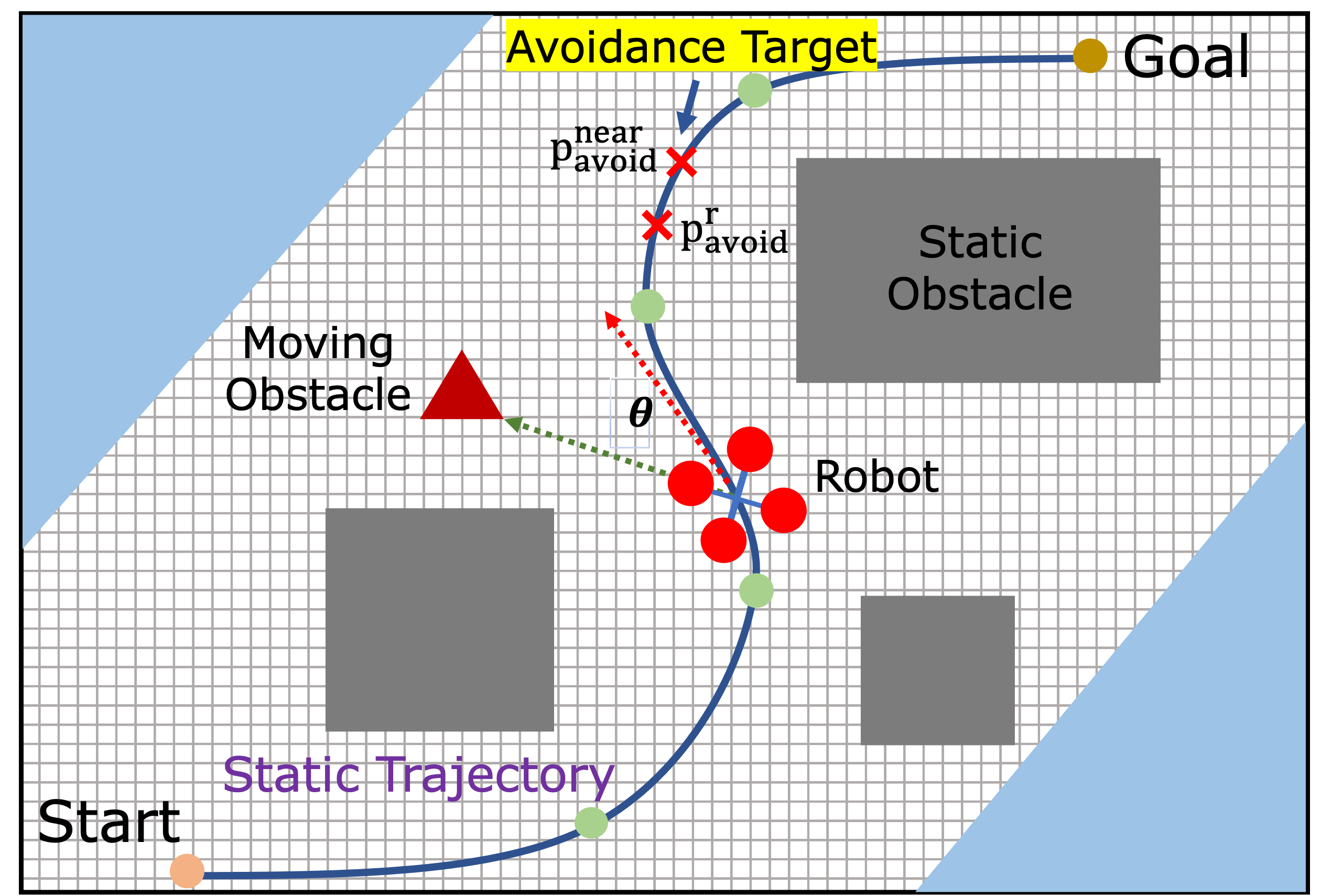}
    \caption{Illustration of avoidance target selection. When the robot meets a moving obstacle, two candidate target positions $\textbf{p}^{\text{near}}_{\text{avoid}}$ and $\textbf{p}^{\text{r}}_{\text{avoid}}$ are selected based on the nearest position from the obstacle on the trajectory and the robot position, respectively. The one with further trajectory distance to the quadrotor is selected as the avoidance target ($\textbf{p}^{\text{near}}_{\text{avoid}}$ is selected in this scenario). Note that $\theta$ of a point is the angle between its moving direction and the obstacle direction.}
    \label{avoidance target}
\end{figure}

The algorithm for finding the avoidance target is presented in Alg. \ref{ATS}. Generally, the algorithm returns the avoidance target from evaluating two avoidance target candidates (Line \ref{candidate1} and Line \ref{candidate2}) based on the nearest position $\textbf{p}^{\text{near}}_{\text{avoid}}$ on the trajectory from the obstacle and the robot position $\textbf{p}^{\text{r}}_{\text{avoid}}$. From our experiment, the two candidates chosen from different perspectives can improve the reliability of the safe detour. From Lines \ref{nearest} - \ref{iteration1}, it first gets position on the static trajectory nearest the moving obstacle $\textbf{p}_{\text{near}}$ from the obstacle and then iterates through the following positions on the trajectory $\sigma_{\text{near}}$ after that point. Once the trajectory distance $\mathcal{D}_{\text{near}}$ from the iterated point to the nearest point is larger than the threshold and the angle $\theta_{\text{near}}$ between the moving direction of iterated point and obstacle direction is greater 90\degree \ (Line \ref{threshold1}), the algorithm stores the first avoidance target $\textbf{p}^{\text{near}}_{\text{avoid}}$ (shown as the upper red cross in Fig. \ref{avoidance target}).  For the second candidate  $\textbf{p}^{\text{r}}_{\text{avoid}}$, we start the iteration from the current robot position (Lines \ref{current} - \ref{iteration2}) and select the point which has the obstacle distance $\mathcal{D}_{o}$ greater than the threshold (Lines \ref{distance2} - \ref{candidate2}) and angle $\theta_{r}$ greater than 90\degree, visualized as the lower red cross in Fig. \ref{avoidance target}. Finally, the avoidance target is the avoidance target candidate whose trajectory distance to the current position is larger (Line \ref{argmaxTarget}).

For the condition of meeting obstacles, we define it based on two requirements: (a) the angle $\theta$ (angle between robot moving direction and obstacle-to-robot direction) as shown in Fig. \ref{avoidance target} is less than 90\degree, and (b) the distance from the robot to the obstacle is less than a predefined threshold. Based on the idea of the angle requirement, we also require the avoidance target always to have $\theta$ greater than 90\degree, which helps the robot drive away from the obstacle.

\begin{algorithm}[t]
\caption{Avoidance Target Selection}
\SetAlgoNoLine%
\label{ATS}
 $\sigma_{\text{traj}} \gets \text{static trajectory}$\;
 $\textbf{p}_{r} \gets \text{robot position}$, $\textbf{p}_{o} \gets \text{obstacle position}$ \;
 $\Delta \gets \text{distance threshold}$ \;

        $\textbf{p}_{\text{near}}$ $\gets $ \textbf{nearestPosInTraj}($\textbf{p}_{o}$, $\sigma_{\text{traj}}$)\; \label{nearest}
        $\sigma_{\text{near}} \gets $ \textbf{trajStartFromPos}($\textbf{p}_{\text{near}}$)\; \label{nearest_traj}
        \For{$\normalfont{\textbf{p}}_{\text{traj}}$ \normalfont{\textbf{in}} \label{iteration1} $\sigma_{\text{near}}$}{ 
        $\mathcal{D}_{\text{near}}$ $\gets$ \textbf{trajDistance}($\textbf{p}_{\text{near}}$, $\textbf{p}_{\text{traj}}$)\;
        \If{$\mathcal{D}_{\normalfont{\text{near}}} \geq \Delta$ \normalfont{\textbf{and}} $\theta_{\text{near},\textbf{p}_{\text{traj}}} \geq \frac{\pi}{2}$}{ \label{threshold1}
                $\textbf{p}^{\text{near}}_{\text{avoid}} \gets \textbf{p}_{\text{traj}}$\Comment*[r]{target candidate 1}\  \label{candidate1}
                \textbf{break}\; \label{break1}
            }
        }
        $\sigma_{\text{r}} \gets $ \textbf{trajStartFromPos}($\textbf{p}_{\text{r}}$)\; \label{current} 
        \For{$\normalfont{\textbf{p}}_{\text{traj}}$ \normalfont{\textbf{in}}  $\sigma_{\text{r}}$}{ \label{iteration2}
        $\mathcal{D}_{\text{o}}$ $\gets$ \textbf{distance}($\textbf{p}_{\text{o}}$, $\textbf{p}_{\text{traj}}$)\; \label{distance2}
        \If{$\normalfont{\mathcal{D}}_{\text{o}} \geq \Delta$ \normalfont{\textbf{and}} $\theta_{\text{o}, \textbf{p}_{\text{traj}}} \geq \frac{\pi}{2}$}{
                $\textbf{p}^{\text{r}}_{\text{avoid}} \gets \textbf{p}_{\text{traj}}$ \Comment*[r]{target candidate 2}\  \label{candidate2} 
                \textbf{break}\; \label{iteration2_exit}
            }
        }
 
 $\textbf{p}_{\text{avoid}} \gets $ \textbf{argmaxTrajDist}($\textbf{p}^{\text{near}}_{\text{avoid}}$, $\textbf{p}^{\text{r}}_{\text{avoid}}$, $\textbf{p}_{r}$)\; \label{argmaxTarget}
 
$\textbf{return} \ \textbf{p}_{\text{avoid}}$\;
\end{algorithm}

\begin{figure*}[t]
    \centering
    \includegraphics[scale=0.65]{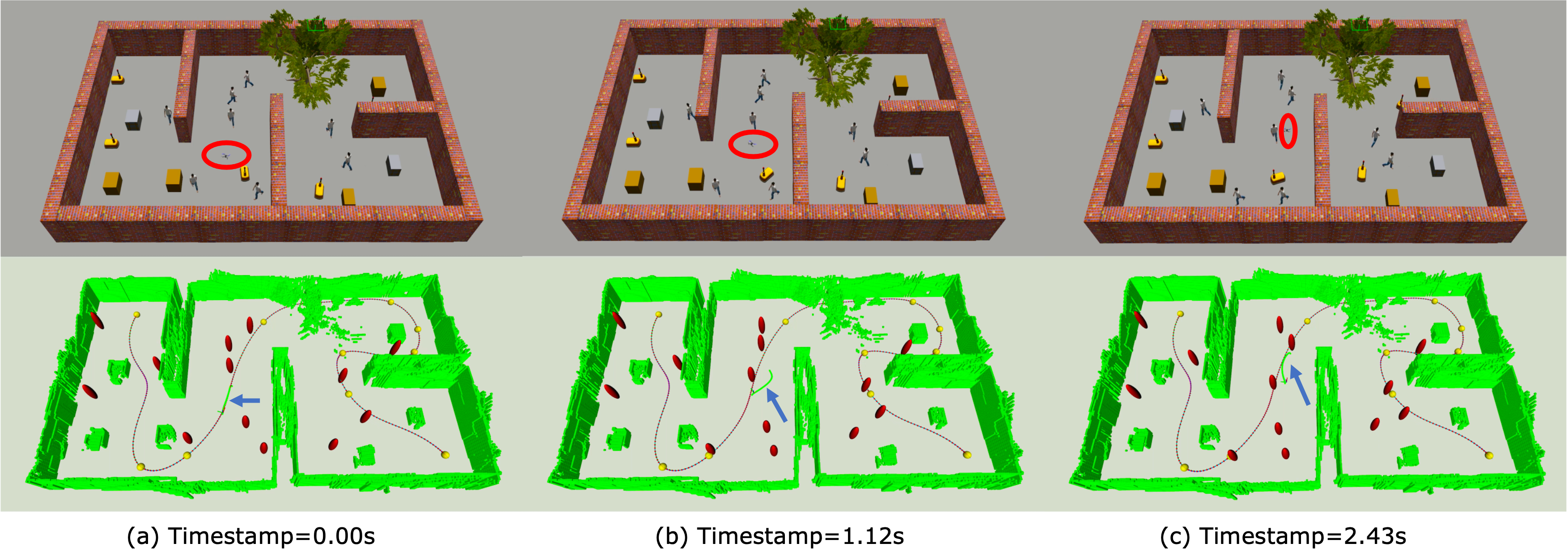}
    \caption{Illustration of dynamic obstacle avoidance in the Maze environment. The quadrotor meets the obstacle at $\text{timestamp}=0.00$s while staying on the static trajectory (red curve). Then at $\text{timestamp}=1.12$s, the chance-constrained MPC takes the temporal to generate a reactive trajectory (green curve) for avoidance. Finally, at $\text{timestamp}=2.43$s, the quadrotor flies away and gets back to the static trajectory again.}
    \label{maze_demo}
\end{figure*}

After obtaining the avoidance target, we can construct the temporal goal for MPC tracking. The temporal goal has the same structure as the static trajectory but only has the length up to the prediction horizon of the MPC. We denote the time step of the avoidance target to be $t_{a}$ on the trajectory, and the target at the $k$th step for the temporal goal is $\sigma^{k}(t_{a})$, which we omit the subscript for static trajectory for simplicity. So, the temporal goal can be written as:
\begin{equation}
\begin{split}
\sigma_{\text{tg}} = [\sigma^{1}(t_{a}),\dots, \sigma^{N-n}(t_{a}), \sigma^{N-n+1}(t_{a +\Delta T}) \\
\dots, \sigma^{N}(t_{a + n\Delta T})],
\end{split}
\end{equation}
where $N$ is the prediction horizon of the MPC. The first repeating $N-n$ terms in the temporal goal are the avoidance target to ensure the robot can avoid the obstacle. The following $n$ terms provide the following static trajectory starting from the avoidance target, which allows a smoother transition from obstacle avoidance to static trajectory tracking. If $n=0$, it indicates no usage of future static trajectory information and will usually increase the difficulties for getting back to the original static trajectory. On the other hand, if $n=N$, there is no repeating avoidance target and will cause the robot not to find a reasonable solution. Based on our experiment, $\frac{n}{N} \in [\frac{1}{4}, \frac{3}{4}]$ is a reasonable setting for both obstacle avoidance and tracking. Finally, the dynamic layer will use the map module to conduct collision checking and output the collision-free part of the reactive trajectory.

\section{Result and Discussion}
\subsection{Experiments Setup}
The experiments are conducted based on PX4 Gazebo simulation environments, which include vehicle dynamics. The algorithm is implemented in ROS with C++ running on Intel Core i7-10750H CPU@2.6GHz. The Octomap \cite{hornung13auro} is used to represent the static map. We apply Quadprog++ \cite{goldfarb1983numerically} for the quadratic programming in the static layer and Acado Toolkit \cite{Houska2011a} for the model predictive control implementation. In our implementation, we choose the polynomial degree to be 7 or 8 in the static planner for both real-time performance and the smoothness of the trajectory. The MPC prediction horizon is set to $N=20$ with 0.1s time step. The obstacle position with its uncertainty is propagated by the linear model: $\textbf{p}^{k+1}_{0} = \textbf{p}^{k}_{0} + \textbf{v}^{k}_{o}(t_{k+1} - t_{k})$ and $\Sigma^{k+1}_{o} = \Sigma^{k}_{o} + \Sigma^{k}_{o,v}(t_{k+1} - t_{k})^2$, where $ \textbf{v}^{k}_{o}$ and $ \Sigma^{k}_{o,v}$ are obstacle velocity and its uncertainty at the $k$th step, respectively.

\begin{table}[h]
\begin{center}
\caption{Environment Information.} \label{env_info}
\begin{tabular}{ c c c  } 
 \hline

 Env. Name & Size ($\text{m}^\text{3}$) & Obstacle Info.\\ 
 \hline

 Maze & $20 \times 10 \times 3$ & 8 persons, 5 robots \\ 

 Tunnel & $15 \times 15 \times 6$ & 6 person, 5 robots \\  

 Forest & $40 \times 40 \times 4$ & 10 persons, 2 robots\\  
 \hline
\end{tabular}
\end{center}
\end{table}

\subsection{Simulation Experiments}
To better evaluate the overall navigation safety, we prepared three simulation environments as shown in Table \ref{env_info}. These environments are on different scales with a different number of dynamic obstacles. All the dynamic obstacles follow the predefined piecewise linear trajectories, and we manually distribute the dynamic obstacles evenly in the environments. The example of the robot avoiding dynamic obstacles is shown in Fig. \ref{maze_demo}. The red ellipsoids represent the dynamic obstacles. And, the red curve represents the static trajectory that the robot should track when it does not meet the dynamic obstacle (Fig. \ref{maze_demo}a). After meeting the obstacle, the temporal goal is generated based on the avoidance target and used as the reference trajectory in our chance-constrained MPC. The green curve denoted by the blue arrow is the generated reactive trajectory for collision avoidance (Fig. \ref{maze_demo}b). Finally, in Fig. \ref{maze_demo}c, the robot goes back to its static trajectory after safely bypassing the dynamic obstacles. The bottom of Fig. \ref{intro_figure} also shows the case of obstacle avoidance when multiple obstacles surround the robot. The complete experiment video is available at \url{https://youtu.be/e03kZ8Zh0AI}.

\begin{table*}[h]
\renewcommand\arraystretch{1.2} 
\begin{center}
\caption{Performance comparison of the proposed DPMPC with and without temporal goal tracking and the deterministic DPMPC under different uncertainty levels in the Forest Environment. The metrics are based on the average minimum distance from obstacles, trajectory length, time,  and success rate.}  \label{comparison_table}
\begin{tabular}{ |c| c c c c| c c c c| c c c c| } 
 \hline
    \multirow{2}{*}{Uncertainty Level} &
      \multicolumn{4}{c|}{0.25$\Sigma$} &
      \multicolumn{4}{c|}{$\Sigma$} &
      \multicolumn{4}{c|}{4$\Sigma$} \\

      & $d_{\text{min}}$ & Length  & Time  & SR   & $d_{\text{min}}$  &  Length  & Time  & SR  & $d_{\text{min}}$  &  Length  & Time  & SR   \\
    \hline

 Proposed DPMPC & 1.66 & 74.51 & 21.67 & 100\% & 1.63 & 75.48 & 24.07 & 100\% & 1.55 & 78.21 & 24.39 & 83.3\% \\

 Deterministic DPMPC & 1.42 & 77.51 & 22.05 & 100\% & 1.35 & 76.71 & 24.43 & 69.2\% & 0.92 & 80.20 & 28.48 & 46.1\% \\

 DPMPC w/o Temporal Goal & 1.65 & 84.30 & 28.89 & 100\% & 1.62 & 87.71 & 29.45 & 100\% & 1.58 & 88.12 & 32.80 & 80.5\% \\
 \hline
\end{tabular}
\end{center}
\end{table*}


\subsection{Analysis and Discussion}

To analyze the performance of the proposed method, we record computational time, obstacle distance, trajectory length, and time in four environments 50 times from different start and goal positions. 

The comparison of the proposed DPMPC planner with its deterministic version and DPMPC without temporal goal tracking is shown in Table \ref{comparison_table}. In deterministic DPMPC, we replace the chance constraint (Eqn. \ref{chance_constrained}) with a deterministic collision constraint without uncertainty consideration. Overall, in all uncertainty levels, the proposed method has the largest average minimum distance from obstacles with the least trajectory length and time and maintains a high success rate.  The deterministic method's success rate drops dramatically due to the increasing uncertainty level, and its distance from obstacles also goes down much in $4\Sigma$ uncertainty. From the data of DPMPC without the temporal goal tracking, we can see that it takes much longer to execute the trajectory compared other two methods. From our observations, although it can avoid the obstacles safely, it is harder to smoothly return to the original static trajectory, which leads to a longer trajectory time. 

\begin{figure}[t]
    \centering
    \includegraphics[scale=0.50]{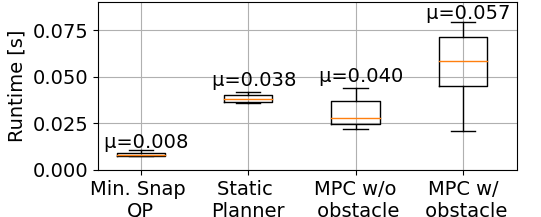}
    \caption{The runtime of each component of our proposed method. When the robot meets obstacles, the computation time of the chance-constrained MPC slightly increases but still can guarantee real-time performance.}
    \label{runtime}
\end{figure}

Fig. \ref{runtime} shows the computational time for both the static and the dynamic layer. For the dynamic layer, we specifically record the runtime of the MPC for both meeting and not meeting dynamic obstacles. Generally, the planner can achieve real-time performance when the robot meets obstacles even though a slight increase in the MPC runtime. The linearized chance constraints largely help improve the optimization efficiency compared to the method mentioned in \cite{blackmore2011chance}. For the static layer, the planner also only requires a small amount of computational time. The efficiency of solving quadratic programming (QP) makes our iterative corridor shrinking algorithm run fast, which usually requires up to 5 or 6 iterations in our experiments. 

\begin{figure}[t]
    \centering
    \includegraphics[scale=0.45]{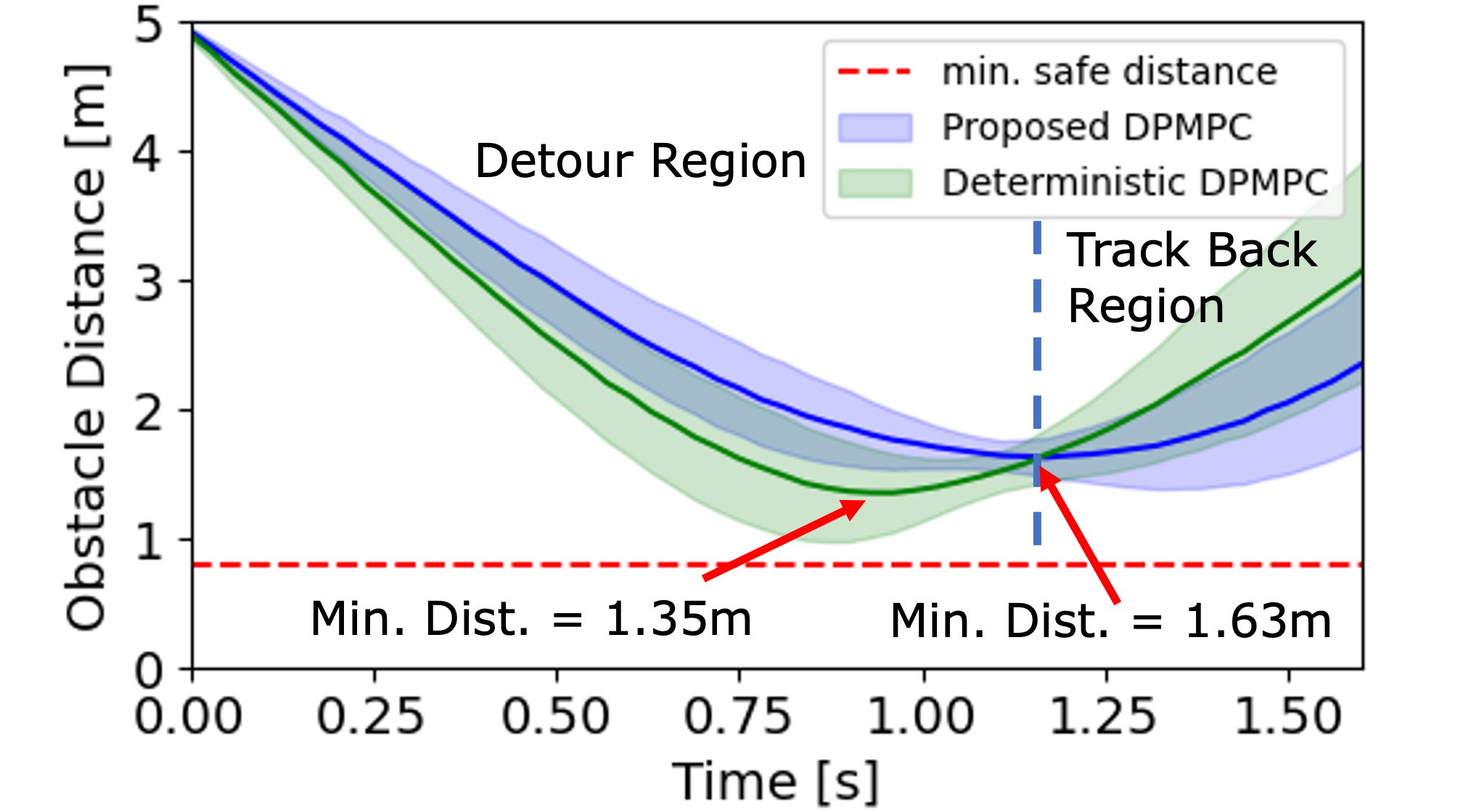}
    \caption{Comparison of the change of the obstacle distance with respect to time when meeting the future colliding obstacle. The proposed method has a larger average minimum distance from obstacles.}
    \label{obstacle_distance}
\end{figure}

\begin{figure}[t]
    \centering
    \includegraphics[scale=0.55]{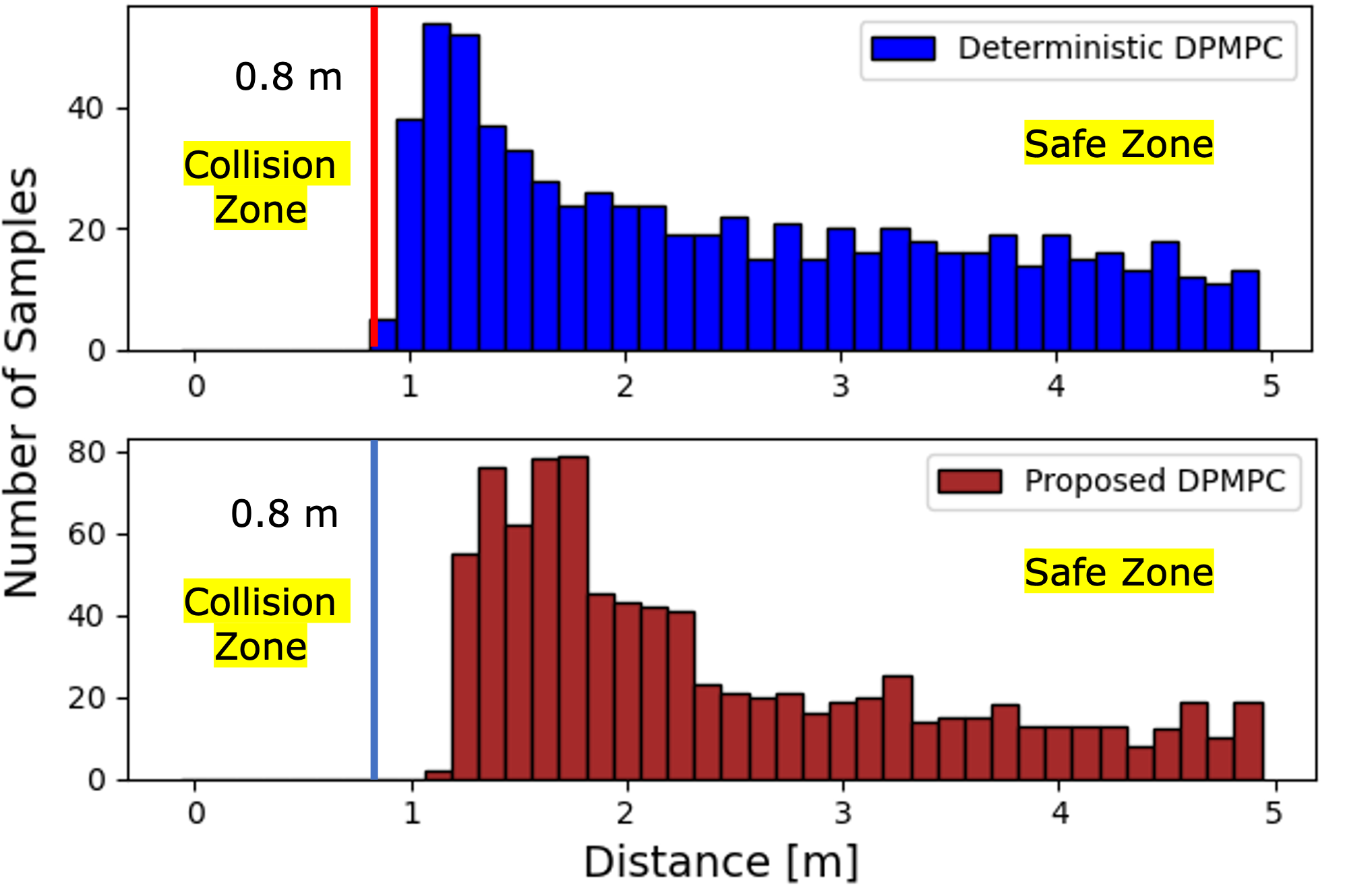}
    \caption{Histogram of distances when meeting obstacles. The proposed method ensures the quadrotor keeps a safe distance from obstacles.}
    \label{distance_bar}
\end{figure}

The behavior comparison of the robot avoiding obstacles can be visualized in Fig. \ref{obstacle_distance}.  The region on the left of the blue dash line shows the process from meeting obstacles to detouring behavior in the proposed DPMPC. After successfully making the detour, the robot will return to its static trajectory and go far away from the obstacle, as shown in the track back region. Our proposed method has a larger average minimum distance from obstacles compared to the deterministic version, and this is also shown in Fig. \ref{distance_bar}, the histogram comparison of obstacle distance, in which the plot of the proposed DPMPC is further to the collision zone. By simply enlarging the obstacles based on uncertainties (e.g. 3 $\sigma$ region) in deterministic DPMPC, we could improve safety performance, but may also make trajectory more conservative since the covairance of obstacle future states grow linearly with respect to time. Compared to deterministic DPMPC, the enlarged obstacle size will become larger when prediction horizon extends.

\section{Conclusion and Future Work}
This paper presents the novel Dynamic Polynomial-based Model Predictive Control (DPMPC) planner for UAV navigation in dynamic environments. Our planner adopts a two-layer scheme fully utilizing the static map and the geometric obstacle representation, allowing the robot to consider complex environment structures and dynamic obstacles. Our iterative corridor shrinking algorithm with quadratic programming (QP) formulation in the static layer helps the robot efficiently generate the collision-free static trajectory. Furthermore, the dynamic layer applies the chance-constrained model predictive control with the temporal tracking method to avoid dynamic obstacles in real-time. The experiment results show the high success rate and the real-time performance under different uncertainty levels. We will implement the whole system in real experiments with onboard vision-based dynamic obstacle detection in our future work.

\section{Acknowledgement}
\noindent The authors would like to thank TOPRISE CO., LTD and Obayashi Corporation for financial support for this work.

\bibliographystyle{IEEEtran}
\bibliography{bibliography.bib}

\end{document}